# Supervised Classification for Object Identification in Urban Areas using Satellite Imagery

Hazrat Ali[1*], Adnan Ali Awan[1], Sanaullah Khan[1], Omer Shafique[1], Atiq ur Rahman[2], Shahid Khan[3]

[1]Department of Electrical Engineering, COMSATS Institute of Information and Technology
Abbottabad, Pakistan
[2]College of Science and Engineering, Hamad Bin Khalifa University, Doha, Qatar
[3]Laboratoire-de-conception-optimisation-et-modélisation-des-systèmes, University de Lorraine, Metz, France
hazratali@ciit.net.pk, anan.malik000@gmail.com, sunnykhan0100@gmail.com, omerawan@gmail.com, arehman@hbku.edu.qa,
engr.shahidkhanshinwari@gmail.com
Correspondence: Hazrat Ali, hazratali@ciit.net.pk

*Abstract*— **This paper presents a useful method to achieve classification in satellite imagery. The approach is based on pixel level study employing various features such as correlation, homogeneity, energy and contrast. In this study gray-scale images are used for training the classification model. For supervised classification, two classification techniques are employed namely the Support Vector Machine (SVM) and the Naïve Bayes. With textural features used for gray-scale images, Naïve Bayes performs better with an overall accuracy of 76% compared to 68% achieved by SVM. The computational time is evaluated while performing the experiment with two different window sizes i.e., 50 × 50 and 70 × 70. The required computational time on a single image is found to be 27 seconds for a window size of 70 × 70 and 45 seconds for a window size of 50 × 50.**

## I. Introduction

Texture study holds a significant role in computer vision tasks [1]. Textural study is widely applied into many research areas including the study of satellite images, industrial quality control, biomedical imagery, and remote sensing. The process of texture judgments can be broadly divided into three phases namely; feature extraction, feature selection and classification/ segmentation. Extraction of texture features deals with the computation of features from the image data such that the features represent information on the spatial distribution of gray level variation in the texture. Generally, a set of features is used for texture discrimination; however, decision on the choice of optimal features is more than often empirical, if not always. The subject of feature selection in texture analysis is concerned with mathematical tools to create the optimal features set characterizing sufficiently the distinguishing properties of the different texture classes.

In this work, high resolution images from the *3D Semantic Labeling Content* as provided by the **International Society for Photogrammetry and Remote Sensing** (ISPRS) are taken as the dataset [2]. These images are rich in textural, contextual, and pattern information and can provide discriminative features.

Features extraction from satellite imagery is achieved through segmentation. Segmentation creates local homogeneity while maintaining global heterogeneity. Based on this understanding, the statistics analysis of all potential scaling parameters can provide a choice of optimal scaling parameter [3].

The grey-level co-occurrence matrix (GLCM) is a prevailing textural feature used to describe spatial properties [4], [5]. In this work, we propose to use GLCM textural features for object identification in aerial imagery. We compare the performance of two classification models namely the Support Vector Machine and the Naïve Bayes trained on the GLCM features. The rest of the paper is organized as follows: The proposed methodology on features extraction and classification is presented in Section II. We present the results obtained on the satellite imagery and the performance comparison of different classifiers in Section III. Finally, we conclude the paper in Section IV.

## II. Proposed Methodology

In this work, we use four GLCM textural features namely *homogeneity, contrast, angular second moment* and *entropy*. The GLCM takes into account both the spectral and the spatial distribution of image grey values. First, we convert the RGB image to gray scale representation. From the gray scale image, the GLCM matrix is calculated. Finally, the four features from GLCM matrix are used to train the classifier. A work flow of the proposed approach is shown in Fig. 1. Binary conversion is used to reduce the complexity and computation time of data. We run the experiment and evaluate the computational time over two different window sizes i.e., [70 × 70] and [50 × 50].

### A. GLCM.

The number of count of intensity values exhibits a specific relationship with other intensity values. Statistics are categorized into first-order, second-order and higher order statistics based on the pixel numbers [4]. Four GLCM textural features were used in this study. Mathematically, these features can be defined as in equations (1) through (4):

*1) Homogeneity*

$$\sum_i \sum_j P_{d,r}(i,j) \qquad (1)$$

*2) Contrast*

$$\sum_i \sum_j [(i,j)^2 P_{d,r}(i,j)] \qquad (2)$$

*3) Energy*

$$\sum_i \sum_j P_{d,r}(i,j)^2 \qquad (3)$$

*4) Entropy*

$$\sum_i \sum_j P_{d,r}(i,j)[-\ln [\sum_j P_{d,r}(i,j)]] \qquad (4)$$

Where *(i,j)* represents the corresponding row, column number, *P (i, j)* is the normalized value in cell *(i, j)*, *d* is the distance separating pairs of pixels; and *r* is the given direction.

These textural features provides good representation of the image, which is then useful for classification tasks in urban land [4].

### B. Image classification

For classification, we use two classification models namely; Naïve Bayes and SVM [5]. A total of 20 images are available in the dataset. The size of each image is of the order of 2700×2700 pixels, and every image contains patches of grass, buildings, roads, and vacant land. The testing samples are different than the training samples and are selected randomly.

### C. Support Vector Machine

The support vector machine is a binary classification algorithm [6]. The implementation idea is that; "*input vectors are non-linearly mapped to a very high-dimension feature space*". In the new feature space obtained, a linear decision boundary is constructured. Generally, SVM offers good generalization ability if an optimal decision boundary is established. We apply SVM for multiclass classification using LIBSVM tool [7].

### D. Naïve Bayes

Naive Bayes has been popular choice for supervised classification tasks as it offers fast and easy implementation [8]. We propose simple, heuristic approach to classify the objects within a given image, using the Naive Bayes classifier. We find that the simple use of the Naïve Bayes provides better classification accuracy as compared to SVM.

## III. RESULTS

Fig. 2 shows an original input image from the dataset. It is a high resolution image (3200 × 2800 pixels), which we pre-process and convert to grayscale representation. Fig. 3 shows the grey scale representation of the image, which we later use for further processing and features extraction. From an image, we segment 960 windows of size 50×50 and 460 windows for 70×70. There are 20 images for training and 20 images for testing in available data set.

Fig. 4 shows the image with window size 70×70, where the image (left) is the in-put image while the image (right) is a visualization of the GLCM matrix. The overall run time for this image with a window size of 70×70 is 27.5 seconds. Fig. 5 shows the image with window size 50×50, where the image (left) is the input image while the image (right) is a visualization of the GLCM matrix. The overall run time for this image with a window size of 50x50 is 42.5 seconds. From Fig. 4 and Fig. 5, it is obvious that the larger window size ([70 × 70]) reduces the required run time. However, there is a compromise on the classification accuracy of the framework using large window size. Fig. 6, 7 and 8 provide visualization for the features extracted from a total of 960 windows of the image. The horizontal axis shows the window number and vertical axis shows the range of features. Table I shows the values of the GLCM features and the corresponding accuracy values for Naïve Bayes and SVM. We calculated accuracy by cross validation of the Naïve Bayes and SVM classifiers on the different window sizes (50×50 and 70×70). Experimental results suggest that Naïve Bayes performs better than SVM on this particular dataset using the four GLCM features.

In Fig. 9, the circled regions show the misclassification between roads, trees and buildings. For future work, we would like to explore deep learning approaches for useful features extraction as used in [9] or use a decision level fusion as used in [10].

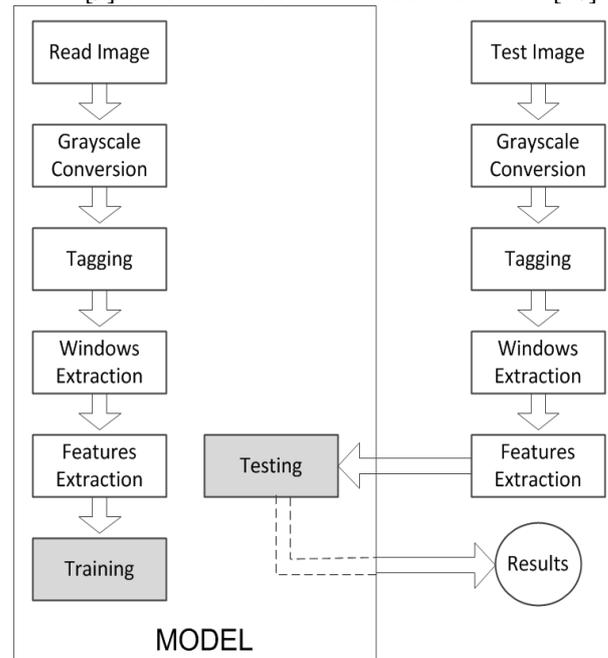

*Fig. 1 Work Flow of the proposed approach*

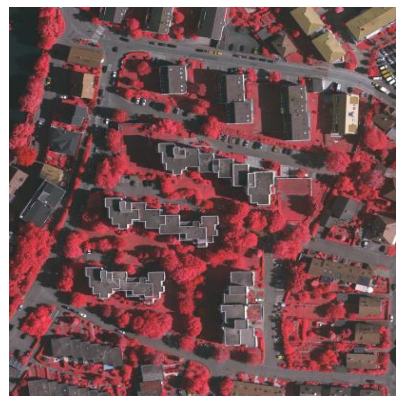

*Fig. 2 An original image from the dataset*

*Table 1 Percentage Accuracy for Naive Bayes and SVM*

| Test Figure | Homogeneity | | Contrast | | Energy | | Entropy | | Naïve Bayes Accuracy | | SVM Accuracy | |
|---|---|---|---|---|---|---|---|---|---|---|---|---|
| | *50×50* | *70×70* | *50×50* | *70×70* | *50×50* | *70×70* | *50×50* | *70×70* | *50×50* | *70×70* | *50×50* | *70×70* |
| 1 | 0.9323 | 0.9310 | 0.9323 | 0.1650 | 0.9323 | 0.4114 | 0.9323 | 0.8413 | 0.9323 | 71.63% | 0.9323 | 64.52% |
| 2 | 0.9409 | 0.9365 | 0.7658 | 0.7632 | 0.5094 | 0.4426 | 0.7501 | 0.08637 | **72.79%** | 71.63% | **67.94%** | 64.52% |
| 3 | 0.9346 | 0.9355 | 0.1550 | 0.1566 | 0.4791 | 0.4145 | 0.7494 | 0.8604 | **70.47%** | 70.37% | **66.92%** | 66.07% |
| 4 | 0.9018 | 0.9317 | 0.1837 | 0.1701 | 0.3951 | 0.3751 | 0.8036 | 0.8630 | **70.37%** | 70.17% | **67.07%** | 66.92% |

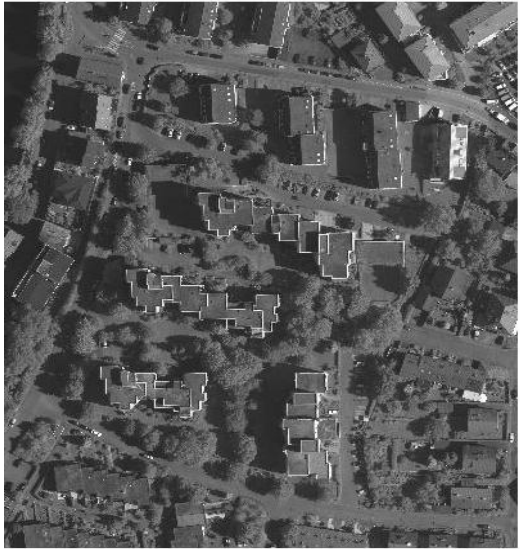

*Fig. 3 RGB to gray scale representation.*

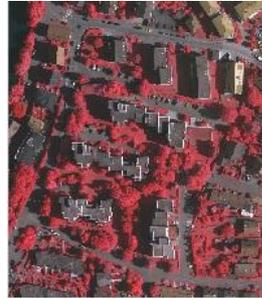 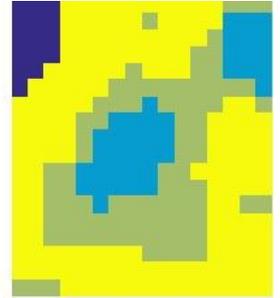

*Fig. 5 A [50 × 50] window size with runtime value in seconds.(Left: Texture Image. Right: Visualization of the GLCM Matrix). Runtime: 42.5 sec*

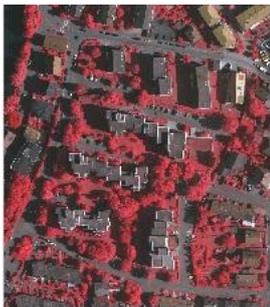 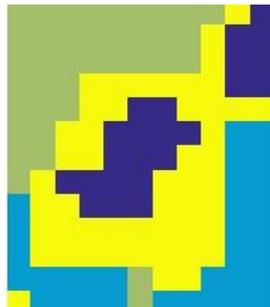

*Fig. 4 A [70 × 70] window size with runtime value in seconds. (Left: Texture Image, Right: Visualization of GLCM Matrix). Runtime: 27.5 sec*

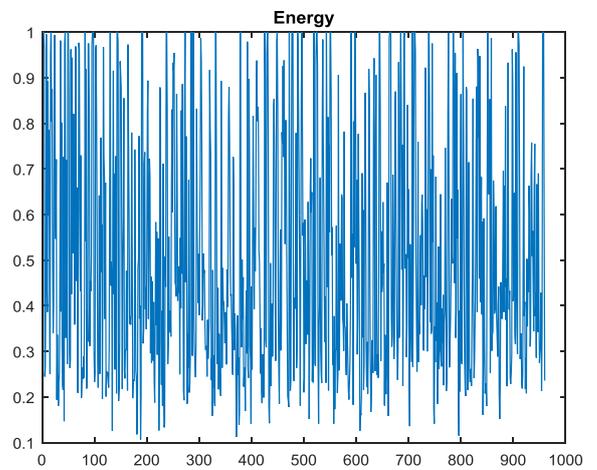

*Fig. 6 Visualization – Energy Features*

.

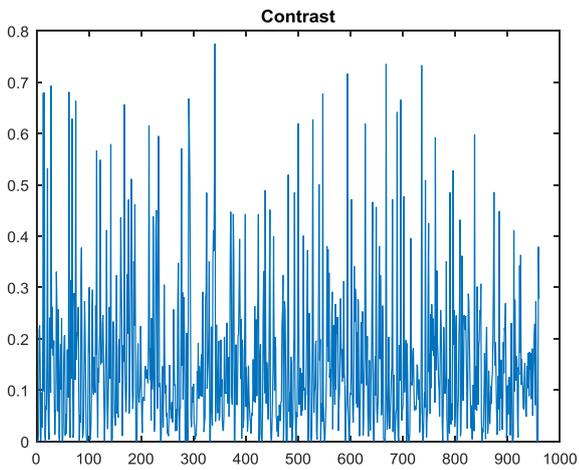

Fig. 7 Visualization – Contrast Features

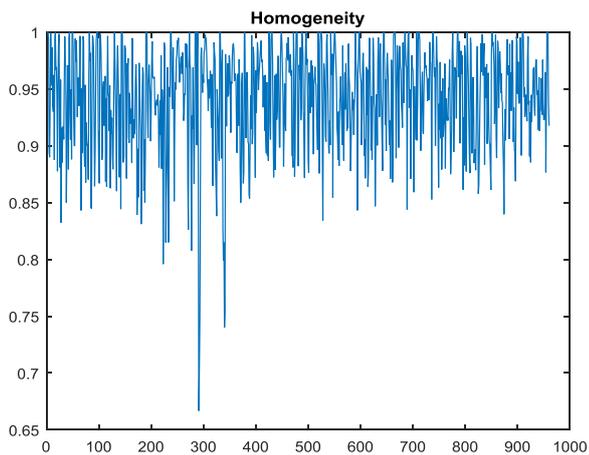

Fig. 8 Visualization – Homogeneity Features

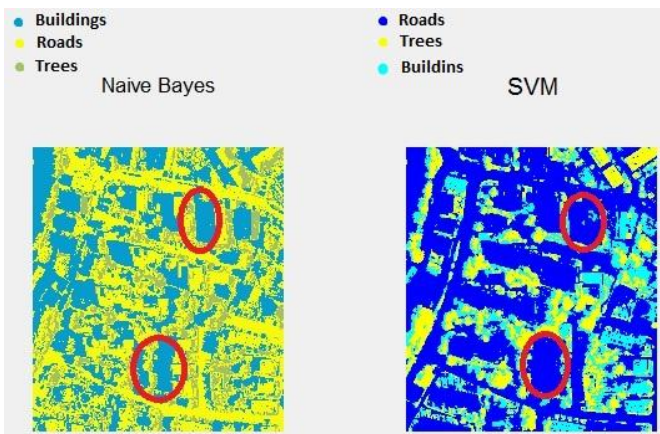

Fig. 9 Misclassification detection

## IV. CONCLUSION

In this work, we have presented an object identification scheme for satellite imagery. We have compared the performance of Naïve Bayes classifier and Support Vector Machine, by training both the classifiers with GLCM features. From Fig. 9 we have observed that SVM misclassify some of the test objects. Consequently, the Naïve Bayes gives better classification accuracy on this task giving an accuracy up to 72% for a 50 × 50 window size. Classification accuracy also varies with the window sizes. Greater the window size, less will be the run time with lower accuracy.